# Barqi breed Sheep Weight Estimation based on Neural Network with Regression

Chintan Bhatt[1], Aboul-ella Hassanien[2], Nirav Alpesh Shah[1], Jaydeep Thik[1]

[1]Charotar University of Science and Technology, Gujarat, INDIA

[2]Scientific Research Group in Egypt, Cairo University, Egypt

**Abstract.** Computer vision is a very powerful method for understanding the contents from the images. We tried to utilize this powerful technology to make the difficult task of estimating sheep weights quick and accurate. It has enabled us to minimize the human involvement in measuring weight of the sheep. We are using a novel approach for segmentation and neural network based regression model for achieving better results for the task of estimating sheep's weight.

**Keywords:** Segmentation, Weight Estimation, Neural Network, Regression

## 1.     Introduction

Sheep's meat as a food source has always been used in high demand. The main factor which affects the proportion of meat of a sheep is the weight of the sheep [1].And thus it becomes a major factor for selecting the sheep. Estimating weight of the sheep through naked eye is not possible for naive consumers because they are not used to such techniques. Also the producers also guess the weight of the sheep by looking at it. Sometimes in some markets and barns sheep weighing is done through weighing scales by placing the sheep on weighing scales one by one which is a very tedious task. So there needs to be a better and optimal way of finding the sheep weight through modern techniques. Computer vision has enhanced over past few years and has made many impossible things possible. Using some advanced computer vision and machine learning algorithms we can find the weight of the sheep remotely without touching the sheep. There are many techniques available to do so and also are feasible techniques [2].

Research with such indirect techniques for weight estimation of many types of animals has already shown accurate results but most of them either require more images or high tech camera with extra functionalities for measuring the linear dimensions. There needs to be way through which a naive consumer can also find the weight of the sheep using mobile device's camera. So we have used a single image taken from normal camera with 3 channels (RGB) to extract the linear dimensions from the image. Also we have included the information about sheep's age and gender to improve the results. Because the gender and age also plays an important role in the shape and weight of the sheep.

Menesatti et al. tried to achieve something similar in [3]. He used two high resolution web cameras connected to the laptop for estimating weight and size of the sheep. This experiment was conducted on 27 sheep of Alpagota type. The height weight and chest depth were measured to estimate sheep's size and weight. The system calculated the linear dimensions by finding the distance between their centers from the two images. Then partial least square regression was used to specify the weight by the distances between points. The efficiency was evaluated by the mean size error. Error was around 3.5% and 5.0% for body length. These were considered as high errors.

Kashiha et al. in [4], used a different approach. The pigs were identified automatically using pattern recognition from the video images. This experiment was conducted on 40 pigs. The video was processed offline and the pigs were recognized from the images using ellipse fitting algorithm which was used to calculate the area of a particular pig. Further this area was used to estimate the weight of the sheep. In this study, pig's weight estimation algorithm achieved 96.2% accuracy.

Prandana et al. in [5], used the side and front image of the cattle for estimation of weight. In this experiment author detected the largest area using active contour and extracted the number of pixels from the image. This was used to estimate the cattle's weight. Front view image was used to detect chest dimensions to add extra features to estimate cattle's weight. Linear regression modeled the relationship among these features and the weight. The accuracy of the system is 73%.

Moreover, in [6], Khojastehkey cropped the image as preprocessing to get the complete side view of new born lambs to estimate their size. Then images were cropped again to get the lamb without head, neck and limbs. Authors depended on that all the lambs are black with white background. Therefore, body was easily recognized by binarizing the image, so cattle size was measured by counting the white pixels. 89% accuracy was achieved in this research.

The main purpose of this paper is to have a real time application which can be used by all type of users. None of the previous approaches can be used to do so because we cannot achieve real time weight estimation as those approaches relied on video recordings and special camera settings. In this work we are using only one side image of the sheep and the age and gender of the sheep to estimate sheep's weight. In this paper dataset of 52 sheep of different physiological conditions and ages has been used. This research has also achieved higher accuracy than its previous approaches of sheep weight estimation. The paper is organized as follows. In Section (2), Theoretical background is explained. In Section (3) proposed weight estimation system is explained. In Section (4) Results achieved through this model is discussed and In Section (5) Conclusion is given.

## 2. Theory and background knowledge

### 2.1 A newer approach to segmentation

Image segmentation is an application of computer vision that can be described as a process that partitions the digital image into segments where each segment represents a different class according to the color scheme. Image segmentation is used here to segment out the sheep from the background in order to estimate the pixel count to determine the surface area of the sheep.

The task of segmentation seems to be difficult here since the haystacks in the background sometimes merges with the texture of sheep wool. This problem has been successfully solved in a newer segmentation approach.

There are various image segmentation algorithms that are available and have been used to solve major segmentation problems in computer vision. The architectures are based on Convolution Neural Nets which is a deep learning approach to solve computer vision. Some of the architectures used are R-CNN [7], Fully Convolution Networks [8] and SegNet [9].

The newer approach is inspired from the SegNet model that uses an encoder-decoder architecture to carry out segmentation task at hand. One thing that is worth noting is that the ground truth in the images used in SegNet architecture is three dimensional, i.e. the three dimensions correspond to the height, width and the number of classes in the image. It uses a per pixel softmax activation at the last layer to determine which class the pixel belongs to.

The newer architecture completely eliminates the task of hand labeling the classes in the third dimension; in fact the third dimension can represent a regular RGB channel. The newer architecture is a auto encoder based architecture with the last layer using a sigmoid activation to determine if a pixel belongs to the sheep or not. This enforces a complete end-to-end training of the segmentation model on the task at hand, thus making it more task specific.

The results of the segmentation are shown in the fig. 1, the input is an RGB image and the output is the segmented sheep.

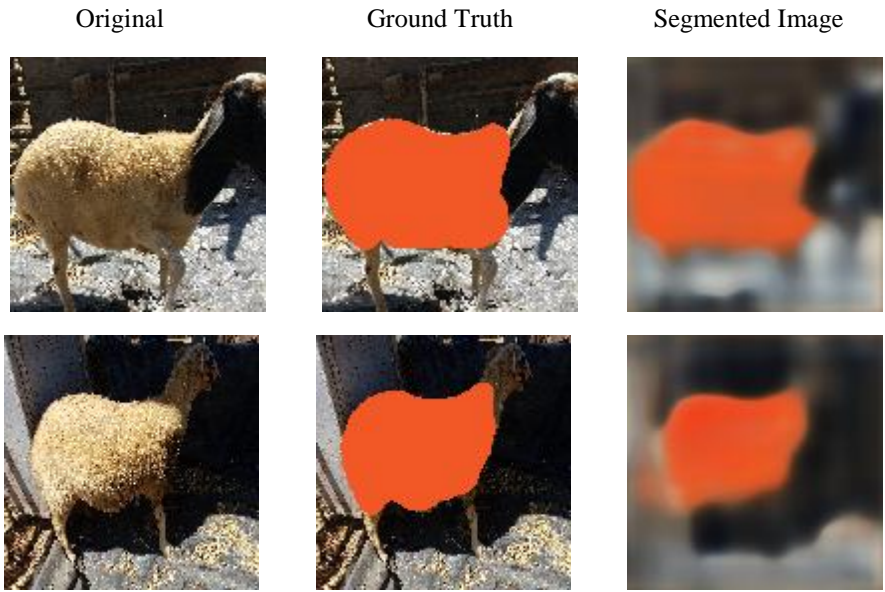

**Fig. 1Segmentation results**

The loss in the final output layer is a binary cross entropy loss given by:

$$H(p,q) = -\sum_x p(x)\log(q(x))$$

Where $p(x)$ is the true underlying Bernoulli distribution and $q(x)$ is the estimated Bernoulli distribution by the learning model.

## 2.2 ANN (Artificial Neural Nets) for regression task

Artificial neural networks are used for clustering through unsupervised learning or classification through supervised learning or regression. That means you can group unlabeled data, categorize labeled data or else predict the continuous values from the historical data.

The regression is mainly used to predict continuous data when a continuous input is provided to it.In Artificial Neural Network X input features are passed from the network's previous layer and are fed to the hidden layer.This X features will be multiplied with the corresponding weights and the sum of that is added to the bias.

$$(X * W) + b$$

This sum is than fed to the rectified linear unit, commonly known as ReLU.It is commonly used because it doesn't saturate on shallow gradients which is the benefit of using ReLU over sigmoid function. For each node ReLU function outputs an activation 'a' which are then multiplied by the corresponding weights ($W_2$) between the hidden and the output layer ( $b_2$ as the bias term) to generate the output . The output layer neuron is a linear function with no activation

$$\hat{y} = a * W_2 + b_2$$

This is the ANN performing regression with single output unit. This result is $\hat{y}$ also known as network's estimate dependent on the X which was supplied as input. This $\hat{y}$ is compared with the $y$ which is also known as ground truth and adjust the weights and biases until the error between them is minimized. This is a simple working of a Artificial Neural Network.

## 3. Proposed Automatic Sheep Weight Estimation System

The main purpose of this paper is to have a real time application that can help any person including buyer or consumer to estimate the sheep weight. And for thus no advanced web cam setting or video recording is required.  Using this technique weight can be estimated by a single photo of the side of the sheep taken from mobile camera and also the age and gender of that sheep which helps to measure the weight of the sheep. The dataset used for this paper includes the images of 52 sheep from a high definition mobile camera. The proposed weight estimation system consists of three phases :(1) Preprocessing phase, (2) Segmentation phase, (3) Weight estimation.

These phases are described in details in the following subsections along with the involved steps with the characteristics for each phase.

### 3.1    Preprocessing Phase

The images are cropped and extra noise from the images is removed manually. Also the images are annotated for training the encoder for segmentation.The sheep in the images are annotated by the orange color and also the original image is saved so as to train the segmentation model. Also the age and gender which are available are organized in a text document separated by space. The annotated cropped image and the annotated image is show in the figure 3.

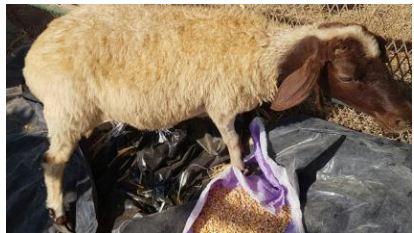 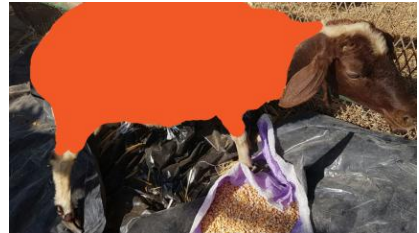

ORIGINAL IMAGE                          ANNOTATED IMAGE

**Fig. 3 Annotated Images**

### 3.2    Segmentation Phase

Segmentation is inevitable part of any computer vision algorithm.There is many algorithms for segmenting and also various ways of doing so. Here we have used unsupervised image segmentation using autoencoder. The detailed discussion of the technique is already done in section 2.1.Using this technique we have segmented the images. The segmented images are further used to calculate the area of the sheep's image using pixel count technique. This area along with other features like sheep's age and gender are used to estimate the sheep's weight.

### 3.3    Estimation Of Sheep's Weight using ANN

Various attempts to predict the weight based on the extracted features using multiple linear regression failed to show up even a decent accuracy on the dataset, this suggested that the underlying function that maps the data to the output weights is not a linear one, and hence the use of neural networks is justified. ANN helped us in modelling a pretty decent approximation of the underlying function.

The structure of the neural net involves a three neuron input layer followed by two hidden layers each with 10 and 5 hidden neurons respectively. The last layer is a single neuron output layer with linear activation to predict the weights. Both the hidden layers have a relu activation units. The ANN is using weight regularization to check overfitting of the model, and the optimizer used is Adam optimizer[10] and mean squared error (MSE) as a loss.

 The MSE is given as

$$MSE = \frac{1}{m}\sum_{i=1}^{m}(y_{pred}-y_i)^2$$

Where $y_i$ is the observed actual value and $y_{pred}$ is the predicted value

The structure of the ANN is as follows

```
Layer (type)                    Output Shape               Param #
=================================================================
dense_1 (Dense)                 (None, 10)                 40

dense_2 (Dense)                 (None, 5)                  55

dense_3 (Dense)                 (None, 1)                  6
=================================================================
Total params: 101
Trainable params: 101
Non-trainable params: 0
```

**Fig. 4 Artificial Neural Network**

### 4. Results and Discussion

We used the data of 52 sheep which included the pixels extracted from the image, the age and gender of the sheep's. We used cross validation to split the data into training and testing sets. We divided the data into 80-20 ratio i.e. 80% Training data and 20% test data. The accuracy of the result was measured using r2 score. r2 score is measured using the below formula

$$R^2 = 1 - \frac{\Sigma(y_i - f_i)^2}{\Sigma(y_i - \bar{y})^2}$$

Where, $y_i$ is the original value and $f_i$ is the predicted value for $i^{th}$ sheep

Using the above metrics, we have achieved a highest r2 score of 0.81 on training set and 0.80 on test set. The results were plotted against the original values on the graph for training set and test set separately which is shown in the fig. 5 and fig. 6 respectively.

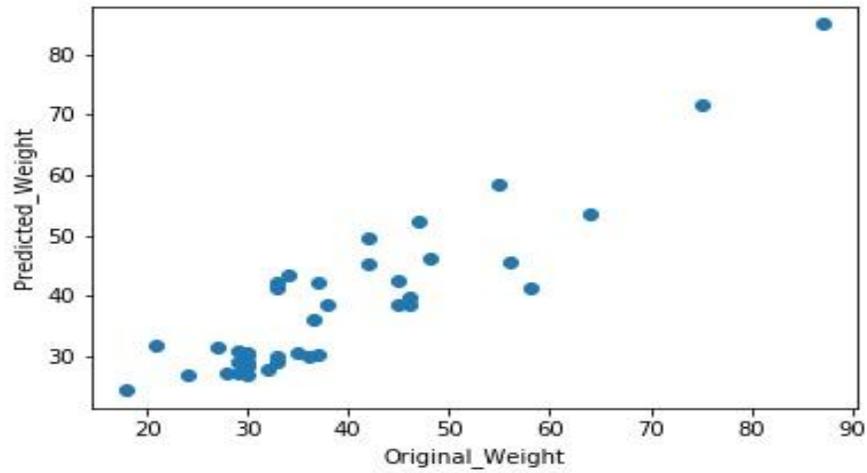

**Fig. 5 Result of training data**

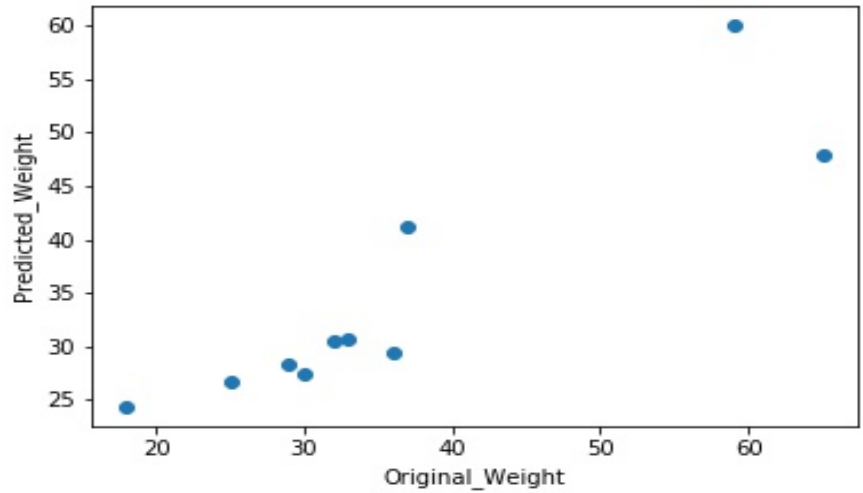

**Fig. 6 Result of test data**

## 5      Conclusions

Using this technique we will be able to estimate the sheep's weight with a simple image from a mobile device. This technique is different than the previous approaches as the previous approaches were either dependent on advanced system setup or high tech cameras. Also the previous approaches didn't take in consideration the age and gender of the sheep which our system takes care of and enables us to get better accuracy than the other similar approaches.

## Acknowledgment

All sheep in this study are from Barqi breed in the 6th of October farm in Ismailia in Egypt. The main origin of this breed is Libya. We thank the staff at the farm to provide these data sets. Refer to [https://arxiv.org/abs/1806.04017] for full description of the sheep images